\documentclass[10pt,conference]{IEEEtran}

\usepackage{multirow}

\usepackage{cite}

\ifCLASSINFOpdf
   \usepackage[pdftex]{graphicx}
   \graphicspath{{figs/}}
   \DeclareGraphicsExtensions{.pdf,.jpeg,.png}
\else
   \usepackage[dvips]{graphicx}
   \graphicspath{{../figs/}}
   \DeclareGraphicsExtensions{.eps}
\fi
\usepackage[cmex10]{amsmath}
\usepackage{amsthm}

\usepackage{algorithmic}

\usepackage{array}

\usepackage{tikz}
\usetikzlibrary{positioning}
\usepackage{dblfloatfix}

\ifCLASSOPTIONcompsoc
  \usepackage[caption=false,font=normalsize,labelfont=sf,textfont=sf]{subfig}
\else
  \usepackage[caption=false,font=footnotesize]{subfig}
\fi
\usepackage{url}

\usepackage{xspace}

\def\onedot{\ifx\@let@token.\else.\null\fi\xspace}
 
\def\ie{\emph{i.e}\onedot}

\def\etal{\emph{et al}\onedot}

\begin{document}
%
\title{Unsupervised representation learning using convolutional and stacked auto-encoders: a domain and cross-domain feature space analysis}

\newif\iffinal
\finaltrue
\newcommand{\jemsid}{77}


\iffinal


\author{\IEEEauthorblockN{Gabriel B. Cavallari, Leonardo S. F. Ribeiro, Moacir A. Ponti}
\IEEEauthorblockA{Institute of Mathematical and Computer Sciences (ICMC)\\
University of S\~ao Paulo (USP), S\~ao Carlos, SP, Brazil\\
Email: [gabriel.cavallari,leonardo.sampaio.ribeiro,ponti]@usp.br}
}


%

\else
  \author{SIBGRAPI paper ID: \jemsid \\ }
\fi

\maketitle

\begin{abstract}
A feature learning task involves training models that are capable of inferring good representations (transformations of the original space) from input data alone. When working with limited or unlabelled data, and also when multiple visual domains are considered, methods that rely on large annotated datasets, such as Convolutional Neural Networks (CNNs), cannot be employed. In this paper we investigate different auto-encoder (AE) architectures, which require no labels, and explore training strategies to learn representations from images. The models are evaluated considering both the reconstruction error of the images and the feature spaces in terms of their discriminative power. We study the role of dense and convolutional layers on the results, as well as the depth and capacity of the networks, since those are shown to affect both the dimensionality reduction and the capability of generalising for different visual domains. Classification results with AE features were as discriminative as pre-trained CNN features. Our findings can be used as guidelines for the design of unsupervised representation learning methods within and across domains.
\end{abstract}


\IEEEpeerreviewmaketitle

\section{Introduction}
\label{s.introduction}

Feature learning is a sub-field of machine learning where trainable models should be capable of inferring good representations (transformations of the original space) from input data alone.  Instead of designing hand-crafted methods to be used in general-purpose scenarios, a model is trained using some dataset so that it learns parameters that are adequate for the data. Deep Learning methods were shown to be effective for the purpose of feature learning, which most recently was defined as representation learning~\cite{Bengio2013representation}. 


For image data, Convolutional Neural Networks (CNNs) with multiple layers were found to be particularly adequate. After being trained for image classification tasks, those network models were shown to be good extractors of low-level (shapes, colour blobs and edges) at the initial layers, and high-level features (textures and semantics) at deeper layers~\cite{Ponti2017}. However, deep networks are difficult to train from scratch, requiring a large number of annotated examples in order to ensure learning, due to their high shattering coefficient~\cite{MelloPonti2018machine}. In the simplest form of feature extraction that require no labels, off-the-shelf pre-trained CNN models already present a good discriminative capacity~\cite{Razavian2014cnn, Dosovitskiy2014discriminative}. CNNs can also be used in a triplet fashion in order to produce a feature embedding for multiple visual domains provided a sufficiently large training set~\cite{Bui2018sketching}.
If given enough network capacity, and enough data, those methods are capable of fitting virtually any labelled dataset, even pure noise sets, highlighting a known issue with CNNs~\cite{Zhang2016understanding}. The generality of feature spaces is then put into question, since small variations in the test set can lead to a significant decrease in testing accuracy~\cite{Nazare2017deep}. This limitation gets worse when some models with enough capacity are able to memorise a dataset, resulting in over-training~\cite{Arpit2017closer} and poor real-life performance. 

Given the aforementioned drawbacks, CNNs are not adequate in two scenarios: a) limited or unlabelled data, and b) when multiple visual domains are considered. In the first case, it is either the event that the dataset is not large enough to allow learning or that the data is not labelled at all, preventing the use of any classification-based training. In the second instance, a convolutional network trained on a given dataset is well fitted to represent it, while the learned feature space may not generalise well for other datasets, especially if it comes from a different visual domain.

In this paper we explore stacked and convolutional auto-encoders as alternative methods for unsupervised feature learning. Auto-encoders (AEs) are methods that encode some input into a representation with low dimensionality, known as \textit{code}, and then, via a decoding module, reconstruct this compact representation to match as best as possible the original input~\cite{Ponti2017} and can be useful in many scenarios, in particular for signal, image and video~\cite{Vega2016single, Shin2013stacked} applications. The space formed by the transformation of the input into the code is often called \textit{latent space}. Encoders and decoders are often linear transformations that can be implemented using a dense layer of a neural network in an unsupervised way~\cite{coates2011analysis}. Stacking those layers may lead to deep stacked auto-encoders that carry some of the interesting properties of deep models~\cite{vincent2010stacked}. 

Based on the idea of self-taught learning that aims to extract relevant features from some input data with little or no annotation~\cite{raina2007self}, we investigate different AE architectures and training strategies to learn representations from images and analyse the feature space in terms of their discriminative capacity.

Although AEs are known to allow unsupervised learning for a specific dataset, usually requiring less examples to converge when compared to CNNs, the ability of those methods to be able to learn latent spaces that can be generalised to other domains, as well as their comparison with features extracted from pre-trained CNNs is still to be investigated. In this paper we explore two datasets, MNIST~\cite{lecun1998mnist} and Fashion~\cite{xiao2017fashion} in order to perform controlled experiments that allow understanding of the potential different AE architectures have in obtaining features that are discriminative not only on the domain of the training dataset, but for other datasets in a cross-domain setting. Note we assume no labels are available for training, hence the unsupervised representation learning. Features obtained with several AE architectures are compared with features extracted from a pre-trained CNN.

\section{Related Work and Contributions}

While Unsupervised Representation Learning is a well-studied topic in the broad field of machine learning and image understanding~\cite{raina2007self}, not much work has been done towards the analysis of those feature spaces when working with cross-domain models. 
The problem we want to tackle by studying the feature space in a cross-domain scenario can be defined as a form of Transfer Learning task~\cite{Pan2010survey}, where one wants knowledge from one known domain to transfer and consequently improve learning tasks within a different domain. 

One of the few studies that leverage the use of auto-encoders in a cross-domain adaptation problem comes from Li et al~\cite{Li2017emotional}; the authors solved their low sample number problem by training an auto-encoder on an unrelated dataset with many samples and using the learned model to extract local features on their target domain; finally, they concatenated those features with a new set learned through usual CNN classifier setup on the desired domain. This study showed the potential for transfer learning with AE architectures but did not experiment with an AE-only design for their models, an application we address in this paper.

Consequently, our contribution includes the following: (i) exploring different AE architectures including dense and convolutional layers, as well as independent and tied-weights training settings (ii) a detailed study using both the reconstruction loss function and the discriminative capability of features extracted using AEs; (iii) an analysis of the feature space on a more extreme cross-domain scenario, in which the images classes are disjoint and with dissimilar visual appearance; (iv) comparison of the discriminative capability of features obtained from AEs and a pre-trained CNN.

To our knowledge, this study is the first that analyses feature spaces of AEs providing guidelines for future work on unsupervised representation learning.

\section{Learning Feature Spaces for Images using Auto-encoders (AEs)}
\label{sec:ae}

\begin{figure}[hpbt]
\begin{center}
    \begin{tikzpicture}[node distance=0.45cm,scale=2, every node/.style={scale=1}]
    \node[minimum size=1cm, name=input, fill=lightgray] (input) {\includegraphics[width=0.125\linewidth]{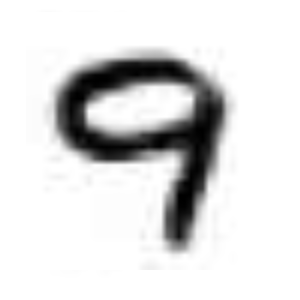}};
    \node[minimum size=1cm, right=of input, fill=red!60!lightgray] (encoder) {Encoder};
    \node[draw,minimum size=1cm, line width=1.5, right=of encoder] (code) {Code};
    \node[minimum size=1cm, right=of code,fill=cyan!40!lightgray] (decoder) {Decoder};
    \node[minimum size=1cm, right=of decoder,fill=gray!75!white] (output) {\includegraphics[width=0.125\linewidth]{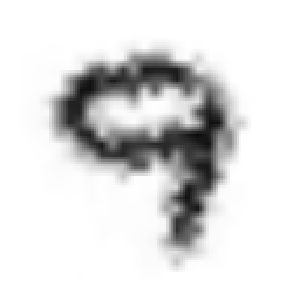}};
    
    \node[above=of input,yshift=-0.5cm] (x) {$x$};
    \node[above=of output,yshift=-0.5cm] (hatx) {$g(z) = \hat x$};
    \node[above=of code,yshift=-0.5cm] (z) {$f(x) = z$};
    
    \draw[->,line width=0.5pt] (input.east) -- (encoder.west);
    \draw[->,line width=0.5pt] (encoder.east) -- (code.west);
    \draw[->,line width=0.5pt] (code.east) -- (decoder.west);
    \draw[->,line width=0.5pt] (decoder.east) -- (output.west);
    \end{tikzpicture}
  \end{center}
  \caption{General structure of AEs.}\label{fig:ae}
\end{figure}
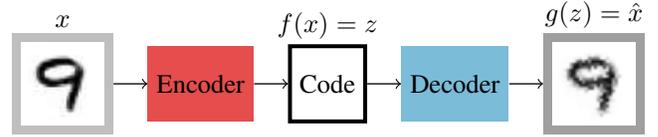

An AE can be implemented as neural network designed to learn an identity function; when an AE takes an example $x$ as input, it should output an $\hat{x}$ that is as similar as possible to the original $x$. From an architectural point of view, an AE can be divided into two parts: an encoder $f$ and a decoder $g$. The encoder takes the original input and creates a restricted representation of it -- we call this representation \textit{code} -- that lies in a space called \textit{latent space}. Then, the decoder is responsible for reconstruct the original input from the code. We illustrate this concept in Figure~\ref{fig:ae}, where the input is an image of a hand-drawn digit. 
 
By restraining the code to be a compact representation of the input data, the AE is forced to learn an encoding transformation that contains the most valuable information about the structure data so that the decoding part is able to perform well in the reconstruction task. The AE cannot however learn to literally copy its input, requiring restrictions to be in place. 


Latent spaces of so-called undercomplete AEs have lower dimensionality than the original input, meaning that its code cannot hold a complete copy of the input data and enforcing that the model should learn how to represent the same data with fewer dimensions. The loss function often used to train an AE is the squared error function:
\begin{equation}
\mathcal{L}(x, g(f(x))) = ||x-g(f(x))||^2
\label{eqn:under_ae_loss}
\end{equation}

\noindent where $L$ is a loss function (e.g. mean squared error), $x$ is an input sample, $f$ represents the encoder, $g$ represents the decoder, $z = f(x)$ is the code generated by the encoder and $\hat x = g(f(x))$ is the reconstructed input data. If functions $f(.)$ and $g(.)$ are linear they can be written in the form:
\begin{align*}
    f(x) &= W_e x + b_e = z\\
    g(z) &= W_d z + b_d = \hat{x},
\end{align*}
were $W_e$ is a weight matrix for the encoder, $b_e$ is a vector of bias terms for the encoder, while the $W_d$ and $b_d$ are, respectively the weight matrix and bias vector for the decoder.

We say the AE has \textit{tied weights} when $W_d = W_e^{t}$, \ie the AE tries to learn a single weight matrix that is assumed to have an inverse transformation given by its transpose.

In summary, we say that an AE generalises well when it understands the data-generating distribution -- \ie it has a low reconstruction error for data generated by such mechanism, while having a high reconstruction error for samples that were not produced by it~\cite{Bengio2013representation}.

Let the decoder $f(.)$ be linear, and $\mathcal{L}(.)$ computed for all training examples be the mean squared error, then the undercomplete AE is able to learn the same subspace as the PCA (Principal Component Analysis), \ie the principal component subspace of the training data~\cite{Ponti2017}. Because of this type of behaviour AEs were often employed for dimensionality reduction. 
Therefore, a side-effect of learning how to reconstruct the input images, the latent space represents a feature embedding for the images that retains the most relevant information about the visual content.

One can build Convolutional AEs by replacing dense layers of a traditional AE with convolutional layers. Those models are useful because they can be designed to obtain hierarchical feature extraction via the auto-encoder architecture. Masci \etal~\cite{Masci11} described convolutional auto-encoders for both unsupervised representation learning and also to initialise weights of CNNs.

\begin{figure}[hpbt]
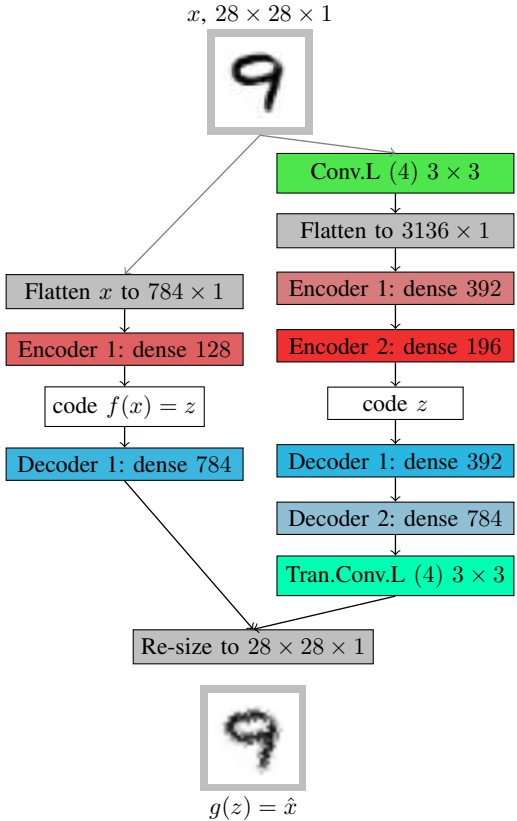

\begin{center}
  \begin{tikzpicture}[node distance=0.85cm,scale=0.75, every node/.style={scale=0.9}]
  
    \node[minimum size=1cm, name=input, fill=lightgray] (input) {\includegraphics[width=0.15\linewidth]{figs/mnist_9.png}};
  
    \node[above=of input, yshift=-1.0cm] (x) {$x$, $28\times28\times1$};
    
    \node [draw,minimum width=3.5cm,below of=input, xshift=-2cm, yshift=-2.25cm,fill=lightgray] (flatten) {Flatten $x$ to $784\times 1$};
    
    \node [draw,minimum width=3.5cm,below of=flatten,fill=red!50!lightgray] (encoder1) {Encoder 1: dense $128$};

    \node [draw,minimum width=2cm,below of=encoder1,fill=white] (code) {code $f(x) = z$};
    
    \node [draw,minimum width=3.5cm,below of=code,fill=cyan!60!lightgray] (decoder1) {Decoder 1: dense $784$};
    
    \node [draw,minimum width=3.5cm,below of=decoder1, xshift=+1.9cm, yshift=-1.85cm, fill=lightgray] (resize) {Re-size to $28\times 28\times1$};

    \node [draw,minimum width=3.5cm,below of=input, xshift=2cm, yshift=-0.5cm,fill=green!60!lightgray] (conv1) {Conv.L $(4)$ $3\times 3$};
    
    \node [draw,minimum width=3.5cm,below of=conv1,fill=lightgray] (cflatten) {Flatten to $3136\times 1$};

    \node [draw,minimum width=3.5cm,below of=cflatten,fill=red!35!lightgray] (cencoder1) {Encoder 1: dense $392$};
    
    \node [draw,minimum width=3.5cm,below of=cencoder1,fill=red!75!lightgray] (cencoder2) {Encoder 2: dense $196$};

    \node [draw,minimum width=2cm,below of=cencoder2,fill=white] (ccode) {code $z$};
    
    \node [draw,minimum width=3.5cm,below of=ccode,fill=cyan!65!lightgray] (cdecoder1) {Decoder 1: dense $392$};
    
    \node [draw,minimum width=3.5cm,below of=cdecoder1,fill=cyan!30!lightgray] (cdecoder2) {Decoder 2: dense $784$};
    
    \node [draw,minimum width=3.5cm,below of=cdecoder2, yshift=0.0cm,fill=green!30!cyan] (conv2) {Tran.Conv.L $(4)$ $3\times 3$};
    
        

    \node[minimum size=1cm, below of=resize, yshift=-0.5cm, fill=lightgray] (output) {\includegraphics[width=0.15\linewidth]{figs/mnist_9_recontruct.png}};
    \node[below=of output, yshift=1.0cm] (hatx) {$g(z) = \hat x$};

    \draw[->,line width=0.5pt,gray] (input.south) -- (flatten.north) node [above,pos=0.5] {};
      
    \draw[->,line width=0.5pt] (flatten.south) -- (encoder1.north) node [above,pos=0.5] {};
    \draw[->,line width=0.5pt] (encoder1.south) -- (code.north) node [above,pos=0.5] {};
    \draw[->,line width=0.5pt] (code.south) -- (decoder1.north) node [above,pos=0.5] {};
    \draw[->,line width=0.5pt] (decoder1.south) -- (resize.north) node [above,pos=0.5] {};
    
    \draw[->,line width=0.5pt,gray] (input.south) -- (conv1.north) node [above,pos=0.5] {};
    \draw[->,line width=0.5pt] (conv1.south) -- (cflatten.north) node [above,pos=0.5] {};
    \draw[->,line width=0.5pt] (cflatten.south) -- (cencoder1.north) node [above,pos=0.5] {};
    \draw[->,line width=0.5pt] (cencoder1.south) -- (cencoder2.north) node [above,pos=0.5] {};
    \draw[->,line width=0.5pt] (cencoder2.south) -- (ccode.north) node [above,pos=0.5] {};
    \draw[->,line width=0.5pt] (ccode.south) -- (cdecoder1.north) node [above,pos=0.5] {};
    \draw[->,line width=0.5pt] (cdecoder1.south) -- (cdecoder2.north) node [above,pos=0.5] {};
    \draw[->,line width=0.5pt] (cdecoder2.south) -- (conv2.north) node [above,pos=0.5] {};
    \draw[->,line width=0.5pt] (conv2.south) -- (resize.north) node [above,pos=0.5] {};
\end{tikzpicture}
\end{center}
 \caption{Illustrations of two undercomplete AE architectures with input $x$ as a $28\times28$ grayscale image. On the left hand side a dense auto-encoder is employed: the image is flattened to a $784\times 1$ vector, then an encoding layer reduce this to a 128-d code, which is then transformed back into a 784-d vector. Finally, the result $\hat x$ is re-sized back to $28\times 28$. On the right hand side, the AE is deeper and starts with a Convolutional layer with 4 filters $3\times 3$ , generating 4 feature maps each with $28\times 28$, which is flattened and given as input to two encoding layers to generate the code. Then three decoding layers (two dense, one transposed convolutional layer) are responsible to reconstruct the image which is re-sized to its original $28\times 28$ size.}
 \label{fig:undercompleteAE}
\end{figure}

To illustrate different architectures, we show examples of undercomplete AEs in Figure~\ref{fig:undercompleteAE}, in which the first has a single layer as encoder and a single layer as decoder so that the code is computed by $z = f(x)$, and the output $\hat x = g(z)$. The second example includes a convolutional layer with 4 feature maps. This has the effect to allow filtering the input image in order to obtain a higher dimensional representation to be  flattened from $28\times28\times4$ to $3136\times1$, obtaining an intermediate representation $x_1 = f_1(x)$. This is then offered to a two-layer encoding process via two nested functions, producing the code $z = f_3(f_2(x_1))$. The code is then transformed by two decoding functions, producing the output as $\hat x = g_2(g_1(z))$.

Each layer usually employ an activation function, allowing the encoding/decoding functions to be non-linear. In this paper we use Rectified Linear Units (ReLU) for Convolutional layers, and Sigmoid functions for dense layers. Note that, by allowing the functions to be nonlinear, and adding several layers we are increasing the capacity of the AE. In those scenarios, despite the fact that their code is smaller than the input, undercomplete AE still can learn how to copy the input, because they are given sufficient capacity. 

In this paper we explore different AE architectures with the objective of learning a latent space that can be used in unseen images for feature extraction with eyes on later retrieval and classification. Note that during the training process no label information is used so this process is fully unsupervised. In the next section we describe the investigated strategies and the experimental setup.

\section{Method}
\label{sec:experiments}

\subsection{Overall experimental setup}

For each dataset, we use the training set only to train the auto-encoders, since we investigate unsupervised learning assuming no labels are available in the training set. The test sets are used to evaluate, via classification accuracy, the feature spaces formed by the auto-encoders as well as the one from a pre-trained Convolutional Network. We do not intent to compare our results with state-of-the art classification methods, but rather to evaluate how discriminative are the representations obtained from the different models.

\subsection{Datasets}

\begin{figure}[h]
\begin{tabular}{c}
\includegraphics[width=0.95\linewidth]{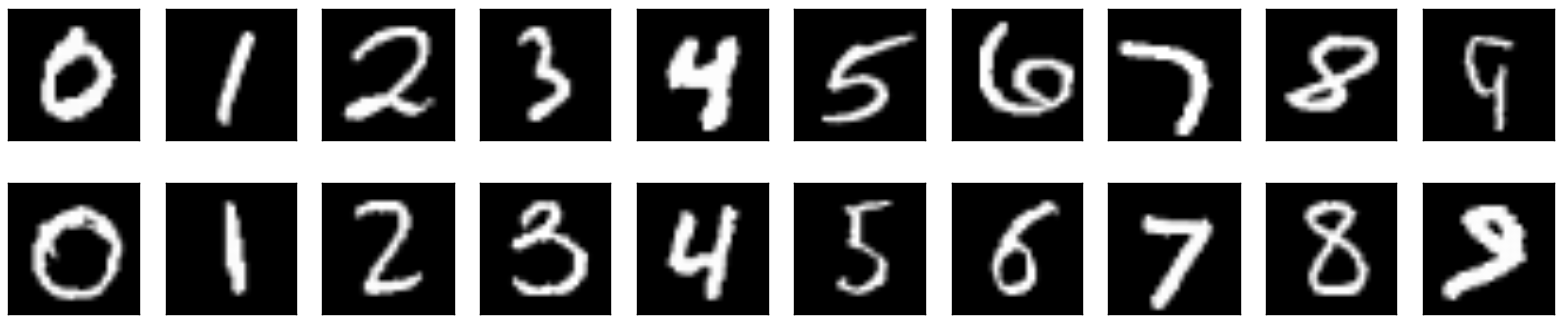} \\
(a) MNIST\\[6pt]
\includegraphics[width=0.95\linewidth]{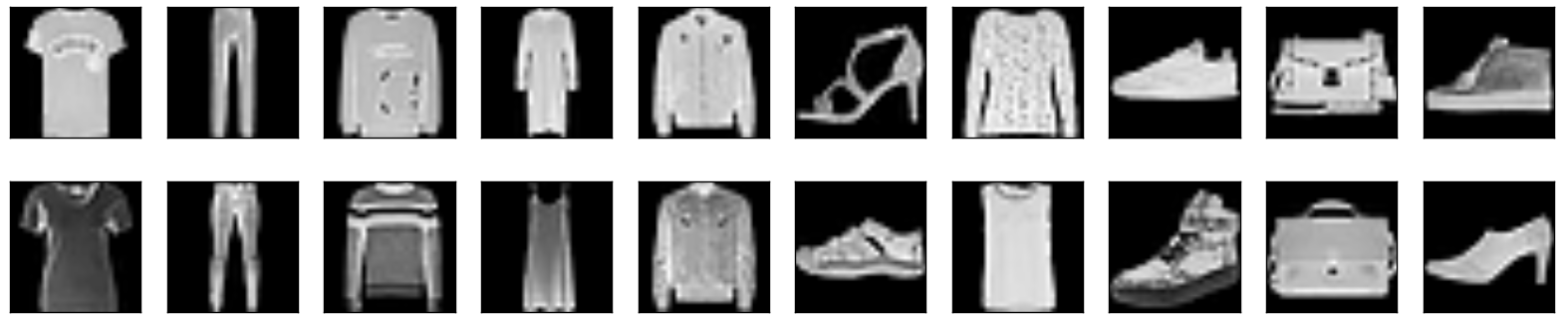}\\ (b) Fashion \end{tabular}
\caption{Examples from the 10 categories of MNIST (a) and Fashion (b)}
\label{fig:dataset}
\end{figure}

The images from MNIST and Fashion datasets were used in our experiments. MNIST has numeric handwritten digits with 60,000 training and 10,000 testing examples~\cite{lecun1998mnist}. The 10 different class of digits (from 0 to 9) are centred in grayscale images with fixed resolution of $28 \times 28$. The Fashion dataset~\cite{xiao2017fashion} was designed to be similar to MNIST in terms of resolution and number of categories, but instead it has clothes and accessories. Figure~\ref{fig:dataset} shows examples of both datasets.

\subsection{AE architectures}

Several architectures and training strategies are investigated, comprising convolutional and dense layers with different sizes. We also vary the size of the code, i.e. the latent space, usually between 256 to 32 dimensions. The different architectures investigated are described as follows:
\begin{itemize}
    \item 2-layer dense (2D): Encoder and decoder with 1 dense layer each, latent space with size 128, 64 or 32: with and without tied weights (total 2 hidden layers);
    \item 4-layer dense (4D): Encoder and decoder with 2 dense layers each, intermediate representation with size 256, latent space with size 128, 64 or 32: with and without tied weights (total 4 hidden layers);
    \item 6-layer dense (6D): Encoder and decoder with 3 dense layers each, intermediate representation with size 392, then 192, latent space with size 128, 64 or 32: with and without tied weights (total 6 hidden layers);
    \item 6-layer dense with conv.layer (6D+C): 1 convolutional layer with 4 filters $3\times 3$, encoder and decoder with 3 dense layers each, intermediate representation with size 392 a latent space with sizes 128, 64 or 32 (total 7 hidden layers);
\end{itemize}

\textbf{Network training:} for all architectures the batch size is 100, with a total of 10,000 iterations on the backpropagation algorithm. A fixed learning rate of $0.0025$ was employed in  the optimisation algorithm RMSprop.

\subsection{Evaluation}

The training set of each dataset is only used to train the auto-encoder. Then the test set is used to compute two evaluation measures:
\begin{itemize}
    \item \textbf{Reconstruction error (AEs):} the first evaluation measure is the mean squared error  (see Equation~\ref{eqn:under_ae_loss}) on the test test images. This is a measure of how well the AE is capable of reconstructing unseen images.
    \item \textbf{Classification accuracy (AEs and CNN):} we employ Support Vector Machine (SVM) classifiers to analyse the linear separability of the latent space, i.e. the space of the codes. Intuitively, a more adequate feature space performs better in SVM, which finds the best as possible linear discriminator with learning guarantees~\cite{Vapnik1999overview, Cortes1995support}. This way the classification accuracy can be seen as a feature space quality measure~\cite{MelloPonti2018machine}. A 10-fold cross validation procedure on the test set is used to obtain mean and standard deviation values of accuracy. This is a proxy measure for the discriminant capability of the latent space for unseen images, \ie images not used to train the auto-encoder or the CNN (we employed a CNN pre-trained using ImageNet as comparison).
\end{itemize}

\section{Results and Discussion}
\label{sec:results}

The results are divided into subsections, in which we analyse: the dimensionality reduction effect, \ie the use of different code sizes. Then we discuss usage of tied weights and convolutional layers. Finally, we show how the AEs behave when trained in one domain but used to reconstruct or extract features from another dataset.

\subsection{Dimensionality Reduction}
\label{ssec:reducedSpace}

\begin{figure*}[h!]
\begin{tabular}{rcc}
   & \includegraphics[width=0.46\linewidth]{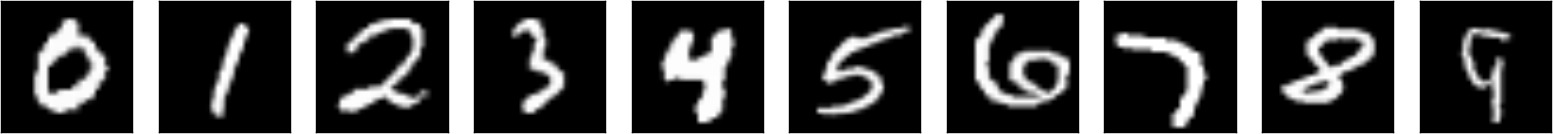} &
\includegraphics[width=0.46\linewidth]{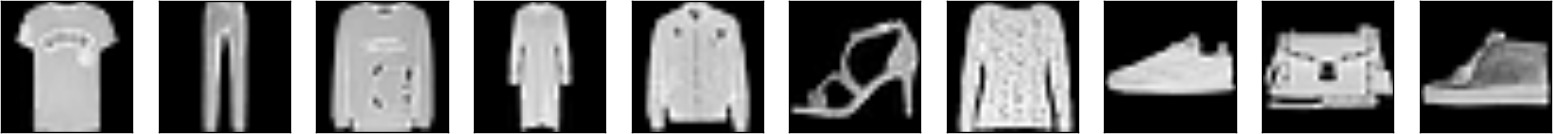} \\
128 & \includegraphics[width=0.46\linewidth]{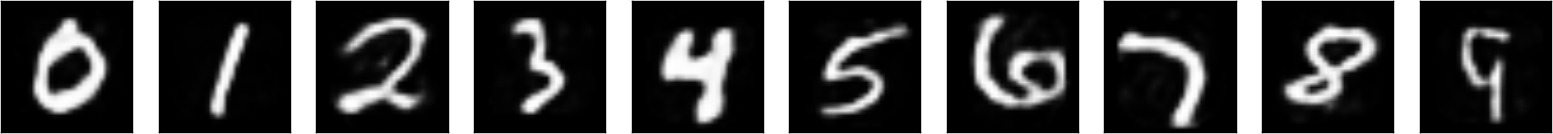} &
\includegraphics[width=0.46\linewidth]{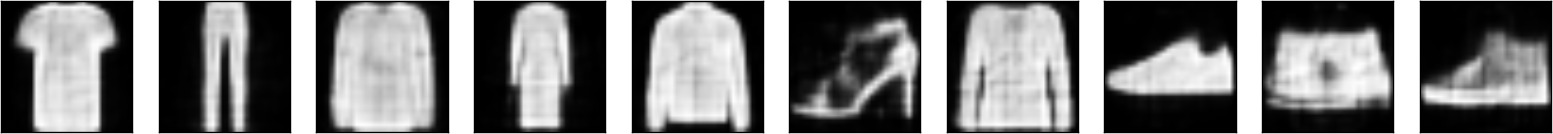} \\
64 & \includegraphics[width=0.46\linewidth]{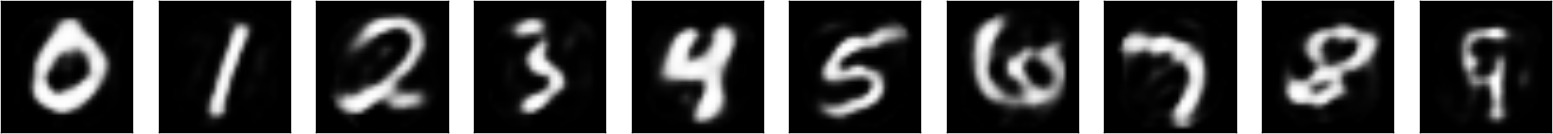} &
\includegraphics[width=0.46\linewidth]{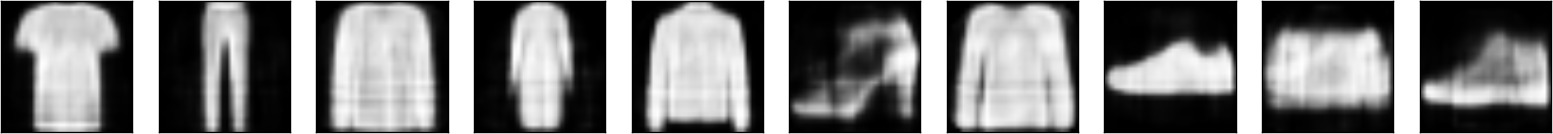} \\
32 & \includegraphics[width=0.46\linewidth]{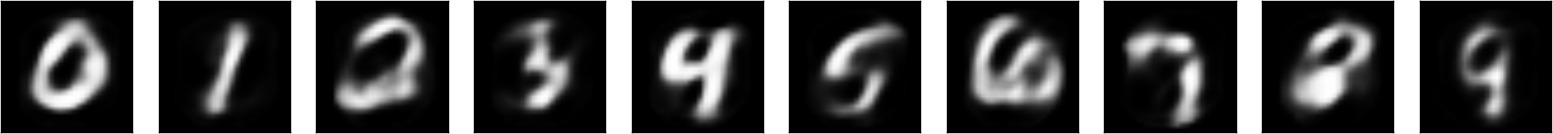} & 
\includegraphics[width=0.46\linewidth]{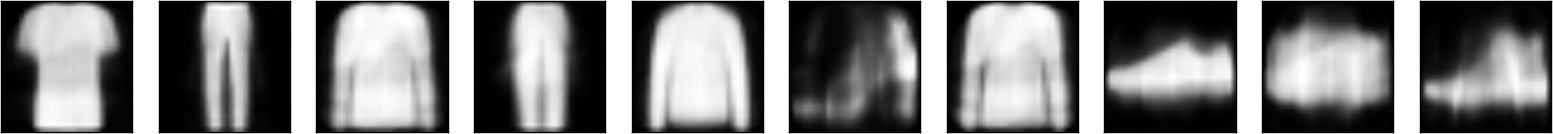}\\
& (a) MNIST & (b) Fashion
\end{tabular}
\caption{Results of AE-2D (2-layer dense) for MNIST and Fashion datasets: first row includes test set images from each category and the remaning rows reconstructions using AE-2D with latent spaces sizes 128, 64 and 32 dimensions}
\label{fig:reduction}
\end{figure*}

The first attempt is to use the simplest architecture possible, with just 2 dense layers (AE-2D). In Figure~\ref{fig:reduction} we show examples of images reconstructed using AE-2D architectures with different code sizes. In Table~\ref{tab:ae2d} the quantitative measures are shown, in which we see the reconstruction error (MSE) increases when a more restricted code is used. For this architecture, the use of tied weights does not look to help. For MNIST, using AE-2D with no tied weighs with a 64 or 128-d code is enough to produce a $92\%$ classification accuracy, while Fashion's best result, $84\%$ was using a 128-d code.

\renewcommand{\arraystretch}{1.2}

\begin{table}[hpbt]
\centering
\caption{Classification results for MNIST using AE-2D architecture}
\label{tab:ae2d}
\begin{tabular}{cc|rrr}
\hline
&    & MSE Train.               & MSE Test   & SVM Accuracy       \\
&& \multicolumn{3}{c}{\em Independent encoder/decoder}            \\
\hline
\parbox[t]{2mm}{\multirow{3}{*}{\rotatebox[origin=c]{90}{MNIST}}} &
32  & $0.02327$                   & $0.02385$ & $0.88 \pm0.06$ \\
& 64  & $0.00941$                   &$0.00878$ & $0.92 \pm0.05$ \\
& 128 & $0.00319$                  &$0.00362$ & $0.92 \pm0.05$ \\
\hline
    && \multicolumn{3}{c}{\em Tied Weights}\\
\hline
\parbox[t]{2mm}{\multirow{3}{*}{\rotatebox[origin=c]{90}{MNIST}}} &
32  & $0.02569$                  &$0.02497$ & $0.88 \pm0.06$ \\
&64  & $0.01530$                    &$0.01533$ & $0.90 \pm0.05$ \\
&128 & $0.00765$                  &$0.00806$ & $0.91 \pm0.05$\\
\hline
\hline
 &   & MSE Train.               & MSE Test   & SVM Accuracy       \\
&& \multicolumn{3}{c}{\em Independent encoder/decoder}            \\
\hline
\parbox[t]{2mm}{\multirow{3}{*}{\rotatebox[origin=c]{90}{Fashion}}} & 32  &$0.02619$ & $0.02578$ & $0.75 \pm0.02$   \\
&64  & $0.01555$ &$0.01637$  &$0.82 \pm0.02$   \\
&128 & $0.00868$ &$0.01002$ & $0.84 \pm0.02$   \\
\hline
 &   & \multicolumn{3}{c}{\em Tied Weights}\\
\hline
\parbox[t]{2mm}{\multirow{3}{*}{\rotatebox[origin=c]{90}{Fashion}}} &32  & $0.02377$ & $0.02565$  & $0.78 \pm0.02$   \\
&64  & $0.01754$ &$0.01746$  & $0.82 \pm0.02$  \\
&128 & $0.01105$ &$0.01161$ & $0.83 \pm0.02$  \\
\end{tabular}
\end{table}

The first question is whether \textbf{increasing the AE capacity allows to produce (i) a more compact code with the same discriminative power and/or (ii) better classification accuracy}. To answer that question we trained a 4-layer dense AE (AE-4D) with and without tied weights. The results are in Table~\ref{tab:ae4d}, showing that it indeed helped on (i) but not in (ii) since dimensionality reduction suffers less in terms of classification accuracy, but overall no significant gain is achieved. However, although we improved MNIST's accuracy by reaching $\sim 94\%$ for 128-d code, which for many applications may be sufficient, the remaining results only improved for smaller codes. In particular, we could not improve the $84\%$ accuracy for Fashion dataset. Since tied weights does not seem to help even on AEs with larger capacity, we decided to not use them in further experiments.

\begin{table}[hpbt]
\centering
\caption{Classification results for MNIST using AE-4D architecture}
\label{tab:ae4d}
\begin{tabular}{cc|rrr}
\hline
&    & MSE Train.               & MSE Test   & SVM Accuracy       \\
&& \multicolumn{3}{c}{\em Independent encoder/decoder}            \\
\hline
\parbox[t]{2mm}{\multirow{3}{*}{\rotatebox[origin=c]{90}{MNIST}}} &
32  & $0.01148$ & $0.01186$ & $0.91 \pm0.04$   \\
&64  & $0.00833$  & $0.00917$  & $0.93 \pm0.04$   \\
&128 & $0.00673$ & $0.00828$ & $0.94 \pm0.04$   \\
\hline
\hline
    && \multicolumn{3}{c}{Tied weights}\\
\hline
\parbox[t]{2mm}{\multirow{3}{*}{\rotatebox[origin=c]{90}{MNIST}}} &
32  & $0.01534$ & $0.01483$ & $0.90 \pm0.06$   \\
&64  & $0.01089$ & $0.01083$ & $0.91 \pm0.05$   \\
&128 & $0.00708$ & $0.00795$ & $0.92 \pm0.04$   \\
\hline
\hline
&    & MSE Train.               & MSE Test   & SVM Accuracy       \\
&& \multicolumn{3}{c}{\em Independent encoder/decoder}            \\
\hline
\parbox[t]{2mm}{\multirow{3}{*}{\rotatebox[origin=c]{90}{Fashion}}} &
32  & $0.01610$ & $0.01671$  & $0.82 \pm0.02$   \\
& 64  & $0.01317$ & $0.01365$  & $0.83 \pm0.02$   \\
& 128 & $0.01200$ & $0.01365$ & $0.84 \pm0.02$   \\
\hline
&& \multicolumn{3}{c}{\em Tied weights}            \\
\hline
\parbox[t]{2mm}{\multirow{3}{*}{\rotatebox[origin=c]{90}{Fashion}}} &
32  & $0.01688$   & $0.01663$ & $0.81 \pm0.02$   \\
&64  & $0.01262$  & $0.01369$ & $0.83 \pm0.02$   \\
&128 & $0.01143$ & $0.01185$ & $0.84 \pm0.01$   \\
\end{tabular}
\end{table} 

\subsection{Convolutional layers}
\label{ssec:convlayers}

\begin{figure*}[hpbt]
\begin{tabular}{rc|c}
& \includegraphics[width=0.46\linewidth]{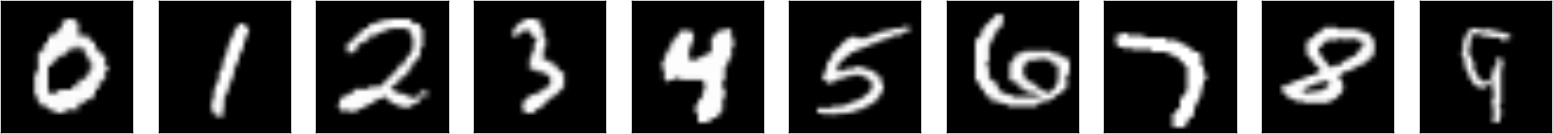} &
\includegraphics[width=0.46\linewidth]{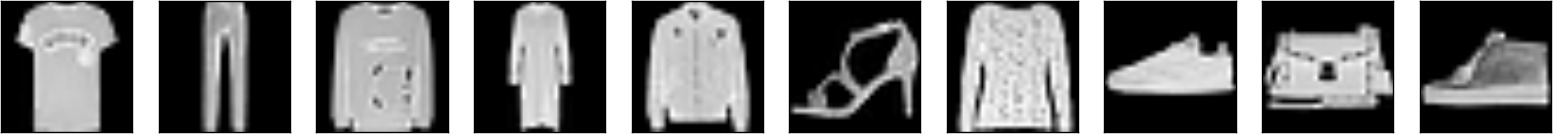} \\
\small 6D &\includegraphics[width=0.46\linewidth]{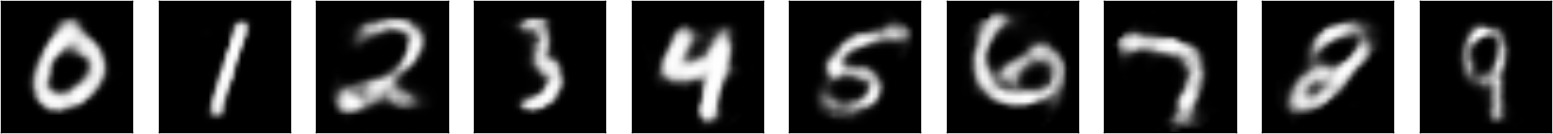} & 
\includegraphics[width=0.46\linewidth]{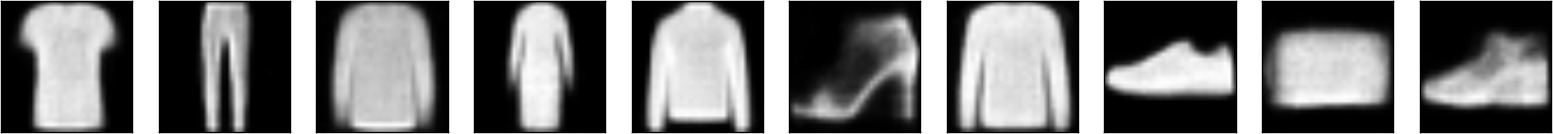}\\
\small 6D+C& \includegraphics[width=0.46\linewidth]{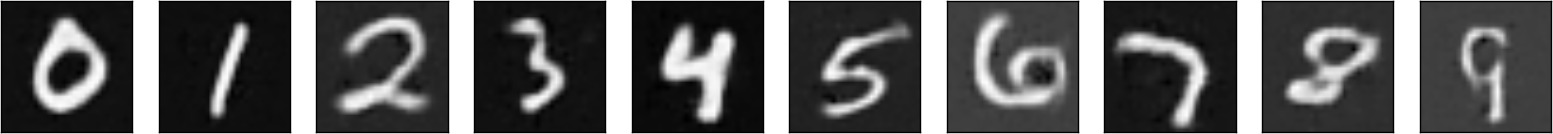} & 
 \includegraphics[width=0.46\linewidth]{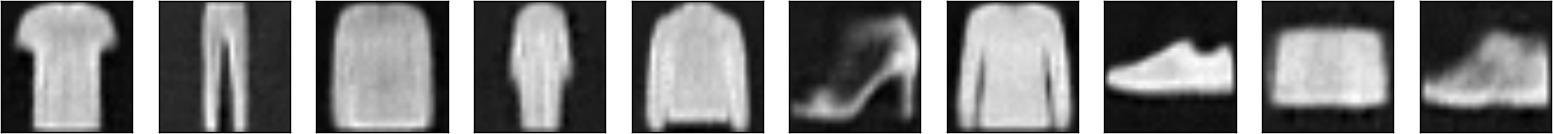} \\
& (a) MNIST & (b) Fashion
\end{tabular}
\caption{Results of AE-6D (6-layers dense) and AE-6D+C (6-layers with a conv.layer) for MNIST and Fashion datasets: first row includes test set images from each category and the remaining rows the AE-6D and AE-6D+C reconstruction results}
\label{fig:ae6d}
\end{figure*}

The second question is related to \textbf{how a convolutional layer would help creating a better feature space}. In order to increase further the capacity of the network we either include a new dense encoding/decoding layer creating AE-6D, or replace one encoding dense layer with a convolutional layer to create AE-6D+C. We experimented with latent spaces with 32, 64, 128 dimensions as before. Figure~\ref{fig:ae6d} shows examples of reconstructed images: in general the convolutional layer seem to have helped reconstructing some details such as the digits 8, 9 for MNIST, as well as for t-shirt, dress and shirt categories of Fashion in terms of the object's grayscale value and similarity with the input image.

\setlength{\tabcolsep}{0.3em} 
\renewcommand{\arraystretch}{1.1}

The classification accuracy results are shown in Table~\ref{tab:ae6d}, in which we see that the convolutional layer was important for learning a discriminative feature space for the datasets, making it possible to use a 32-d code for both datasets with small decrease in the accuracy. 

\setlength{\tabcolsep}{0.5em} 

\begin{table}[hpbt]
\centering
\caption{Classification results for MNIST using AE-6D architecture}
\label{tab:ae6d}
\begin{tabular}{cc|rrr}
\hline
&    & MSE Train.               & MSE Test   & SVM Accuracy       \\
&& \multicolumn{3}{c}{\em Dense (AE-6D)}            \\
\hline
\parbox[t]{2mm}{\multirow{3}{*}{\rotatebox[origin=c]{90}{MNIST}}} & 
  32  & $0.02127$ & $0.02254$ & $0.92 \pm0.04$   \\
&64  & $0.01639$ & $0.01973$ & $0.94 \pm0.04$   \\
&128 & $0.01728$ & $0.01674$ & $0.95 \pm0.03$   \\
\hline
    && \multicolumn{3}{c}{\em Convolutional}\\
\hline
\parbox[t]{2mm}{\multirow{3}{*}{\rotatebox[origin=c]{90}{MNIST}}} &
32  & $0.01284$  & $0.01288$ & $0.93 \pm0.04$   \\
&64  & $0.00898$  & $0.00981$ & $0.93 \pm0.05$   \\
&128 & $0.00977$ & $0.01012$ & $0.94 \pm0.04$   \\
\hline
\hline
&    & MSE Train.               & MSE Test   & SVM Accuracy       \\
    && \multicolumn{3}{c}{\em Dense (AE-6D)}            \\
\hline
\parbox[t]{2mm}{\multirow{3}{*}{\rotatebox[origin=c]{90}{Fashion}}} & 
32  & $0.02063$ & $0.02108$ & $0.78 \pm0.03$   \\
&64  & $0.01922$ & $0.02051$  & $0.78 \pm0.02$   \\
&128 & $0.01826$ & $0.01873$ & $0.80 \pm0.02$   \\
\hline
&& \multicolumn{3}{c}{\em Convolutional}\\

\hline
\parbox[t]{2mm}{\multirow{3}{*}{\rotatebox[origin=c]{90}{Fashion}}} & 
32  & $0.01349$ & $0.01654$ & $0.85 \pm0.02$   \\
&64  & $0.01755$ & $0.01604$ & $0.87 \pm0.03$   \\
&128 & $0.01189$ & $0.01345$ & $0.88 \pm0.02$   \\
\end{tabular}
\end{table}

\subsection{Cross-domain analysis}
\label{ssec:crossdomain}
In order to complement this analysis, we perform a cross-domain experiment in which one dataset is used as ``source'' to train the AE, which is then used to construct the feature space of another ``target'' dataset, for which only the test images are used. 

\begin{figure*}[hpbt]

\begin{tabular}{rc|c}
& \includegraphics[width=0.46\linewidth]{figs/mnist_1c_original.jpg} & \includegraphics[width=0.46\linewidth]{figs/fashion_1c_original.jpg}
 \\
 \small 4D & \includegraphics[width=0.46\linewidth]{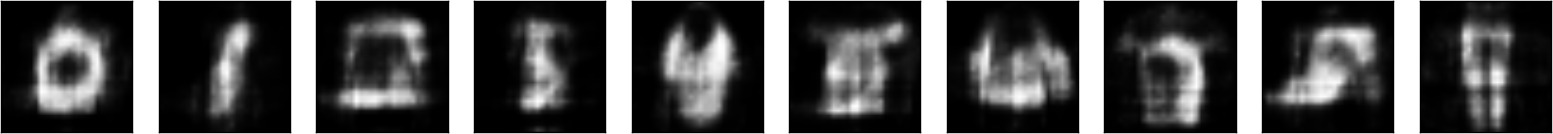} & \includegraphics[width=0.46\linewidth]{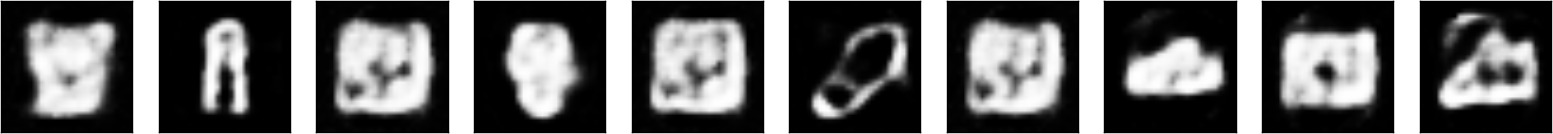}\\ 
 \small 4D-TW & \includegraphics[width=0.46\linewidth]{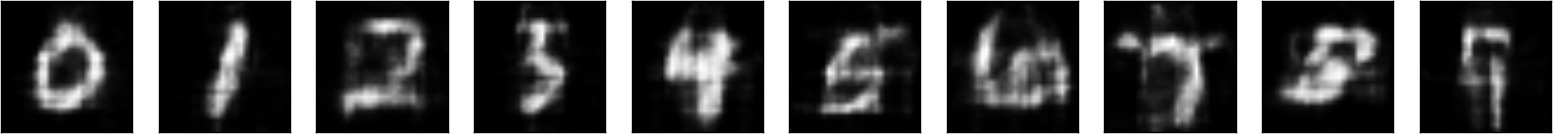} & 
 \includegraphics[width=0.46\linewidth]{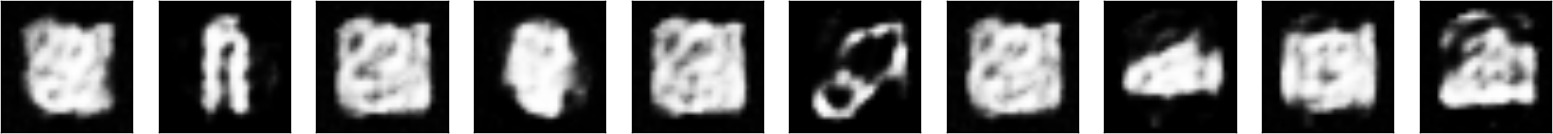}\\
 \small 6D & \includegraphics[width=0.46\linewidth]{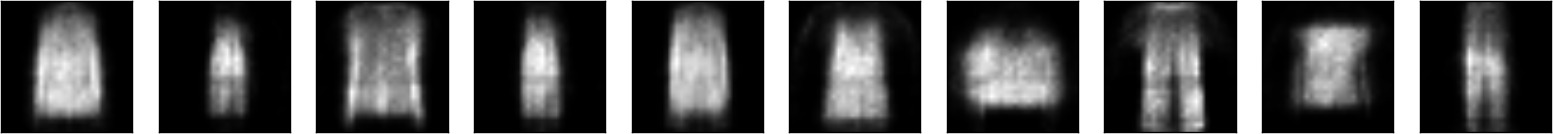} & \includegraphics[width=0.46\linewidth]{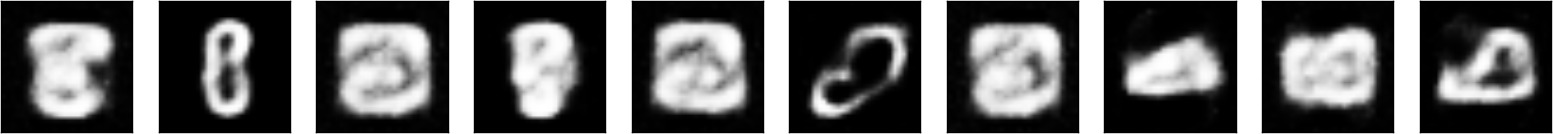} \\
 \small 6D+C   & \includegraphics[width=0.46\linewidth]{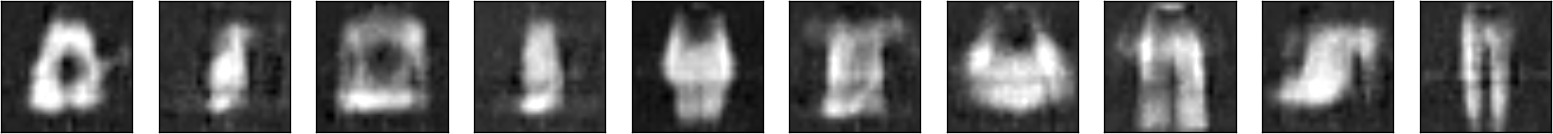} & \includegraphics[width=0.46\linewidth]{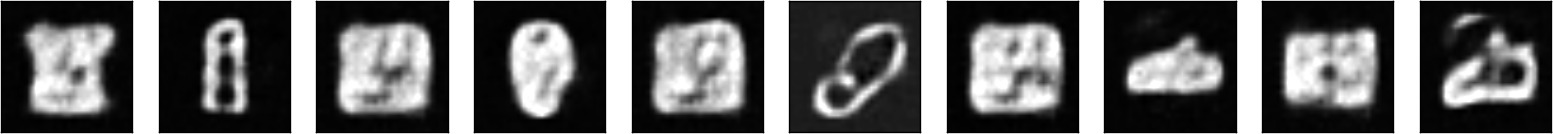} \\
& (a) Fashion (training) to MNIST (testing) & (b) MNIST (training) to Fashion (testing)
\end{tabular}
\caption{Results of cross-domain experiments with AE-4D, AE-4D-TW (tied weights), AE-6D (6-layers dense) and AE-6D+C (6-layers with a conv.layer) for MNIST and Fashion datasets: (a) training on Fashion and using the AE model to reconstruct images and obtain the MNIST representation, (b) training on MNIST and using the AE model to reconstruct images and obtain the Fashion representation. The first row are the original images, and the ones below the reconstructed versions considering the different AE architectures.}
\label{fig:cdae}
\end{figure*}

Thus, in this section we try to answer a third question about learning features: \textbf{how well an AE trained in some dataset is useful to other domains}? This allows to understand if we can rely on weights that are learned considering a specific domain in order to obtain features that are still discriminative for images with a different visual content. In order to answer this question, two scenarios were considered:
\begin{itemize}
    \item Fashion source (training set), MNIST target (test set): image reconstruction results in Figure~\ref{fig:cdae}-(a), and quantitative results in Table~\ref{tab:mnist-fashion},
    \item MNIST source (training set), Fashion target (test set): image reconstruction results in Figure~\ref{fig:cdae}-(b), and quantitative results in Table~\ref{tab:fashion-mnist},
\end{itemize}

\setlength{\tabcolsep}{0.3em} 
\renewcommand{\arraystretch}{1.2}

\setlength{\tabcolsep}{0.5em} 

\begin{table}[hpbt]
\centering
\caption{Results of Cross-Domain Fashion to MNIST}
\label{tab:fashion-mnist}
\begin{tabular}{l|rrr}
\hline
& \multicolumn{3}{c}{Encoder/decoder independentes}            \\
    & MSE Train.               & MSE Test   & SVM Accuracy        \\
\hline
4D        & $0.00904$ & $0.03498$ & $0.94 \pm0.04$ \\
4D-TW     & $0.00776$ & $0.02651$ & $0.94 \pm0.03$    \\
6D        & $0.01423$ & $0.05904$ & $0.80 \pm0.06$ \\
6D+C      & $0.01298$ & $0.04406$ & $0.93 \pm0.04$    \\
\end{tabular}
\end{table} 

\begin{table}[hpbt]
\centering
\caption{Results of Cross-Domain MNIST to Fashion}
\label{tab:mnist-fashion}
\begin{tabular}{l|rrr}
\hline
    & MSE Train.               & MSE Test   & SVM Accuracy \\
\hline
4D    & $0.00527$ & $0.07631$ & $0.82 \pm0.02$ \\
4D-TW & $0.00414$ & $0.08374$ & $0.82 \pm0.02$ \\
6D    & $0.01079$ & $0.08630$ & $0.78 \pm0.02$ \\
6D+C  & $0.00946$ & $0.08772$ & $0.83 \pm0.02$ \\
\end{tabular}
\end{table} 

The quantitative results indicate that an AE with less capacity usually performs better considering both image reconstruction and classification accuracy. This corroborates the intuition stated in~\cite{goodfellow2016deep} that says an AE with enough capacity could learn a one-dimensional code such that every similar instance is mapped to a single neuron in the bottleneck layer using the encoder. This may be what happens when inspecting AE-6D reconstructions from Fashion to MNIST (see Figure~\ref{fig:cdae}(a)), in which the digits are reconstructed to resemble clothes, and also from MNIST to Fashion (see Figure~\ref{fig:cdae}(b)) in which for example the trousers (second image from left to right) is reconstructed by the AE-6D almost as an '8' digit.

Even with imperfect reconstruction, Fashion dataset can be used successfully to train an AE that serves as an MNIST feature extraction, achieving $94\%$ accuracy. As a comparison, supervised CNNs trained with the whole 60k MNIST images often produce around $98\%$ accuracy on test set, while complex semi-supervised approaches that learn with 3k labels produce $95-98\%$ accuracy~\cite{Kingma2014semi}. Considering we learn the embedding using a different dataset, a $0.94\%$ accuracy indicates the potential of unsupervised representation learning.

The opposite way (from MNIST to Fashion) is more challenging: this is in fact expected since MNIST lacks object with larger flat regions and texture, as well as different grayscales. As a comparison, deep methods can achieve $87-91\%$ accuracy on Fashion dataset~\cite{Rieger2017separable, xiao2017fashion}. Even using the limited MNIST as training images, the latent space is still sufficient to allow some degree of linear separability between the classes, achieving $83\%$ accuracy.

\subsection{Unsupervised representations: AE versus pre-trained CNN}
\label{ssec:AEvsCNN}

Since many studies employ pre-trained CNNs as a way to obtain features often using the ImageNet weights, we performed a comparison between the discriminative capability of features: obtained from an auto-encoder and obtained from the feature maps of the last convolutional layer of a ResNet50~\cite{He2016} that was trained with ImageNet. This allows to shed some light on the question: \textbf{how well does off-the-shelf CNN features perform when compared with the AE features?} The results are shown in Table~\ref{tab:CNNvsAE} indicating that off-the-shelf CNN features are no better than those obtained via AEs, in particular those composed of both convolutional and dense layers. For the Fashion dataset, in particular, the classification accuracy of AE features are slightly higher when compared with those computed with ResNet50.

\begin{table}[hpbt]
\centering
\caption{Comparison between the AE (with convolutional layer, 128 features), AE cross domain (with convolutional layer, 256 features) and pre-trained CNN (ResNet50, 2048 features).}
\label{tab:CNNvsAE}
\begin{tabular}{cc|r}
\hline
&    & SVM Accuracy       \\
\hline
\parbox[t]{2mm}{\multirow{3}{*}{\rotatebox[origin=c]{90}{MNIST}}} 
&AE-Conv-128    & $0.94 \pm0.04$   \\
&AE-CD-Conv-256 & $0.93 \pm0.04$   \\
&ResNet50-2048  & $0.96 \pm0.03$   \\
\hline
\hline
&    & SVM Accuracy       \\
\hline
\parbox[t]{2mm}{\multirow{3}{*}{\rotatebox[origin=c]{90}{Fashion}}}  
&AE-Conv-128    & $0.88 \pm0.02$  \\
&AE-CD-Conv-256 & $0.83 \pm0.02$   \\
&ResNet50-2048  & $0.83 \pm0.02$   \\
\end{tabular}
\end{table}

\section{Conclusion}
\label{sec:majhead}

In this paper we show how AEs can be used to obtain features and use those features across different visual domains. First, the dimensionality of the code to be learned is dependent on the capacity of the AE: a deeper AE seems to allow a more compact latent space keeping the accuracy. Second, the inclusion of convolutional layers allow a better overall result in terms of both error of reconstruction and linear separability of the feature space. The third conclusion is related to the cross-domain experiments, in which the use of tied weights helped achieving a lower MSE, also reflecting on a better reconstruction of the images. With respect to the cross domain feature extraction, one must either keep the AE with a limited capacity, or include a convolutional layer to help filtering spatial relationships between the pixels. Finally, those features are comparable or slightly better when compared with the features obtained with a state-of-the art CNN.

The use of AEs is convenient because it allows unsupervised representation learning with a good degree of linear separability, with potential to be transferred across different domains. Future work might investigate the use of transfer learning methods in order to improve cross-domain feature embedding, as well as exploring other flavours of AEs, such as the denoising, contractive and regularised versions.

\iffinal
\section{Acknowledgments}

The authors would like to thank FAPESP (grants \#2016/16111-4, \#2017/22366-8) and CNPq (Researcher Fellowship grant \#307973/2017-4) for financial support. This work is also partially supported by the CEPID-CeMEAI (FAPESP grant \#2013/07375-0).
\fi

\bibliographystyle{IEEEtran}
\bibliography{example}
\end{document}